\documentclass[runningheads]{llncs}

\usepackage{eccv}

\usepackage{eccvabbrv}

\usepackage{graphicx}
\usepackage{booktabs}
\usepackage{color}
\usepackage{multirow}

\usepackage{caption}
\usepackage{booktabs}

\usepackage{arydshln} 

\usepackage[accsupp]{axessibility}  

\newcommand{\mymodel}{COMET}

\usepackage{hyperref}

\usepackage{orcidlink}

\begin{document}


\title{Comprehensive Attribution: Inherently Explainable Vision Model with Feature Detector}

\titlerunning{Comprehensive Attribution for Vision Models}

\author{Xianren Zhang\inst{1}\orcidlink{0000-0003-0283-4693} \and
Dongwon Lee\inst{1}\orcidlink{0000-0001-8371-7629} \and
Suhang Wang \inst{1}\orcidlink{0000-0003-3448-4878}}

\authorrunning{X. Zhang et al.}


\institute{The Pennsylvania State University, University Park, PA, USA \\
\email{\{xzz5508,dongwon,szw494\}@psu.edu}}

\maketitle

\begin{abstract}
As deep vision models' popularity rapidly increases, there is a growing emphasis on explanations for model predictions. The inherently explainable attribution method aims to enhance the understanding of model behavior by identifying the important regions in images that significantly contribute to predictions. It is achieved by cooperatively training a selector (generating an attribution map to identify important features) and a predictor (making predictions using the identified features). Despite many advancements, existing methods suffer from the incompleteness problem, where discriminative features are masked out, and the interlocking problem, where the non-optimized selector initially selects noise, causing the predictor to fit on this noise and perpetuate the cycle. To address these problems, we introduce a new objective that discourages the presence of discriminative features in the masked-out regions thus enhancing the comprehensiveness of feature selection. A pre-trained detector is introduced to detect discriminative features in the masked-out region. If the selector selects noise instead of discriminative features, the detector can observe and break the interlocking situation by penalizing the selector. Extensive experiments show that our model makes accurate predictions with higher accuracy than the regular black-box model, and produces attribution maps with high feature coverage, localization ability, fidelity and robustness. Our code will be available at \href{https://github.com/Zood123/COMET}{https://github.com/Zood123/COMET}.



\keywords{Machine Learning \and  Explainability \and Feature Attribtion}
\end{abstract}

\section{Introduction}

Deep vision models have shown great performances across various Computer Vision tasks \cite{redmon2016you,deng2009imagenet,redmon2018yolov3}. A critical requirement for deep vision models is the interpretability of their predictions. This is essential for experts and lay-users without the expertise to understand and trust the model. Moreover, interpretability can be imperative in high-stakes scenarios such as healthcare \cite{holzinger2017we} and legal regulations \cite{house2022blueprint,doshi2007accountability,regulation2018general}. Feature attribution, which identifies important input features that impact a model's prediction, can be understood by both experts and laypeople \cite{akhtar2023survey}. Feature attribution method typically generates attribution maps with the size of the input image, where each element of the map indicates the importance of the corresponding pixel. There are two lines of attribution approaches, i.e., {\em post-hoc} and {\em inherently explainable} method \cite{zhou2016learning,selvaraju2017grad,srinivas2019full,petsiuk2021black}. First, post-hoc methods aim to provide explanation for a pre-trained and fixed model. Generally, they aggregate information of the input image, activations \cite{zhou2016learning,selvaraju2017grad} and gradients \cite{srinivas2019full} of the pre-trained model, and the output logits \cite{petsiuk2018rise} to generate attribution maps. However, post-hoc approaches can not reduce the opacity of the pre-trained model where the decision process is still not revealed. Moreover, it has been shown that post-hoc attribution methods can be unfaithful, as the attribution map may not accurately reflect the true importance of the input features in making a prediction \cite{jethani2021have,srinivasrethinking}. To reveal how models use available features and guarantee the faithfulness of explanation, second, inherently explainable methods build models by jointly training a selector and a predictor \cite{chen2018learning,jethani2021have,ganjdanesh2022interpretations}. Specifically, the selector first generates an attribution map that selects features and the predictor only adopts the selected features to make predictions. In this way, the feature attributions are more faithful because the prediction is only made by features selected by the attribution map. 

\begin{figure*}[t]
\centering
\begin{minipage}[c]{0.52 \textwidth} 
    \includegraphics[width=\linewidth]{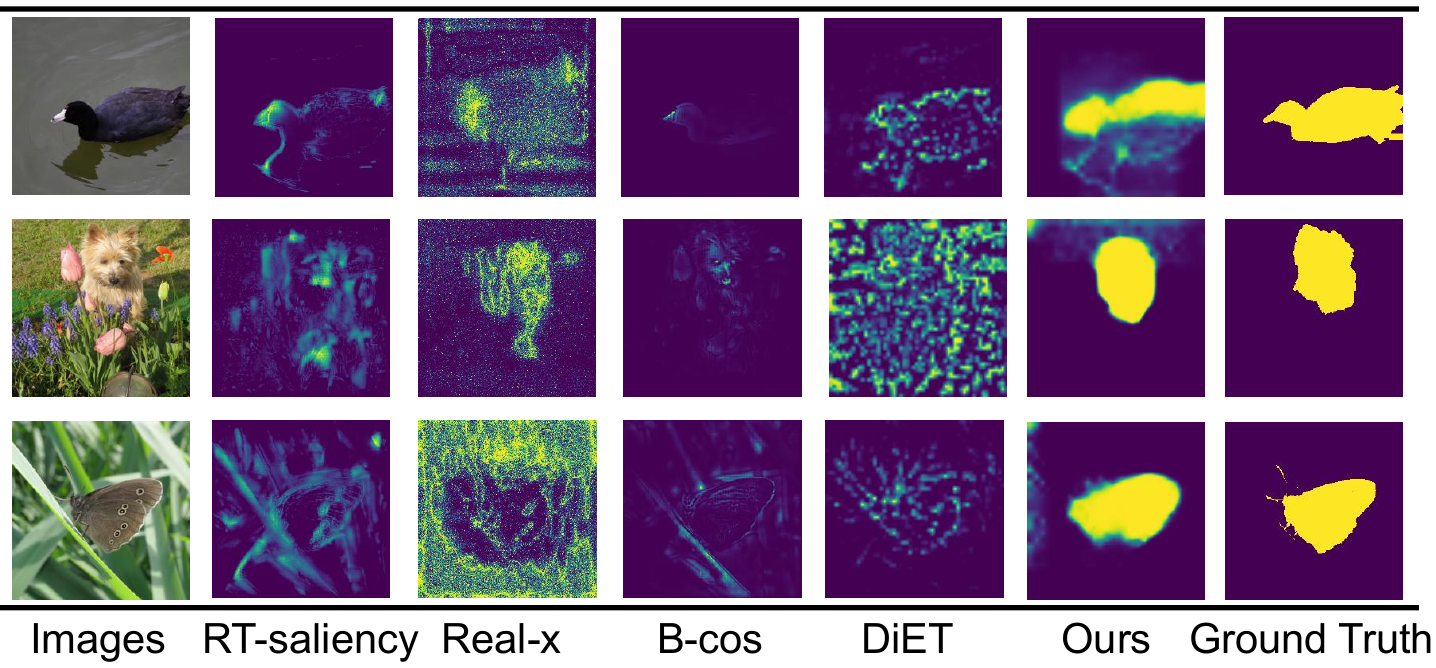}
    \caption{Attribution maps from different explanation approaches. }
    \label{fig: maps_gallory}
\end{minipage}  
  \hfill
  \begin{minipage}[c]{0.46\textwidth}
    \small
\captionof{table}{The Cross-Entropy (CE) loss when the predictor fits on the foreground(F.) or background(B.). The experiment is conducted on ImageNet-9 with ResNet18 as the predictor.}   

\begin{tabular}{ c c c c}
\cline{1-4}
\multirow{2}{*}{CE Loss}& \multicolumn{2}{c}{Predictor} &  \\
& fit on F.        & fit on B.        &  \\ 
\cline{1-4}
Foreground input & \textbf{0.0027} & 0.0298 &  \\
Background input & 0.0320 & \textbf{0.0080} &  \\ 
\cline{1-4}

\end{tabular}
    \label{tab: interlocking}
  \end{minipage}
\end{figure*}

However, existing inherently explainable methods have the {\em incompleteness problem} where not all discriminative features are highlighted. As shown in \cref{fig: maps_gallory}, existing explanation techniques tend to cover only parts of the target. For example, B-cos \cite{bohle2022b} focuses primarily on the beak of the duck, overlooking other parts. A recent study also shows that models tend to exploit shortcuts and only use 5\% of input pixels for predictions \cite{carter2021overinterpretation}. In high-stakes scenarios, such as medical imaging, completeness is crucial for decision-making. The doctor may need the attribution map to know which part of the CT scan indicates the cancer result. The attribution map should contain all abnormal areas (e.g., tumors) to avoid misdiagnosis. The incompleteness problem arises from the current objective which only maximizes the prediction score. The objective encourages the selector to select features that are sufficient enough to support prediction (sufficiency) but not necessarily include all discriminative features (completeness). 

Aside from the incompleteness problem, recent works \cite{jethani2021have,ganjdanesh2022interpretations,yu2021understanding} observe that joint training of the selector and the predictor tends to yield {\em degenerate cases} where the selector selects noise, resulting in low prediction accuracy. This issue arises from the interlocking problem of the selector and predictor \cite{yu2021understanding}. Initially, the non-optimized selector would select noise and mask out discriminative features. The predictor then fits on the noise resulting in poor prediction accuracy. Conversely, the selector is updated to minimize the prediction loss by gradients that flow through the predictor. If the predictor has fitted on noise, the selector will be reinforced to preferentially choose the noise for low prediction loss. We also conduct an experiment on ImageNet9 to demonstrate this problem. The selector can select foreground or background. Similarly, the predictor can either fit on the foreground or background. Then, we evaluate the Cross-Entropy loss when the selector and the predictor use different strategies shown in \cref{tab: interlocking}. The strategy of selecting and fitting on the foreground (Foreground input, fit on F.) yields the globally minimal loss, which aligns with our goal for the model to focus on discriminative features. However, selecting and fitting on the background (Background input, fit on B.) has lower loss than the other two strategies. This suggests that the model could easily fall into a sub-optimal interlocking state, selecting noise over discriminative features. The ground truth labels are the foreground objects and we assume that there are no spurious features in the background, which should consist solely of noise.

To address the aforementioned problems, we propose a novel inherently explainable model called  COMprehensive fEature aTtribution (\textbf{\mymodel}). For the incompleteness problem, we enhance the comprehensiveness of feature selection with a new objective for masked-out regions. Alongside the existing objectives, we introduce an additional objective that discourages the presence of discriminative features in the masked-out regions. This means the selector is penalized if any discriminative features are not selected. This approach ensures a comprehensive and accurate attribution map, addressing the incompleteness problem effectively. For the interlocking problem, considering that the predictor can be misled and fit on the noise, we introduce a feature detector. This detector is designed to detect any discriminative features in the masked-out regions. If the selector mistakenly masks out discriminative features, concentrating instead on less important ones (noise), the predictor will fit on this noise, led astray by the selector. In such cases, the detector can identify these overlooked discriminative features and penalize the selector. The detector can break the interlocking between the selector and the predictor, avoiding degenerate cases. Additionally, our framework has several applications such as improving trustworthiness, helping model debugging \cite{adebayo2020debugging,rao2023using}, and image generation \cite{rombach2022high}. Detailed discussions and examples are in the supplementary materials. 

Our \textbf{main contributions} are as follows: (i) We introduce a novel inherently explainable model that improves feature selection comprehensiveness by discouraging the exclusion of discriminative features. Our method has more comprehensive attribution maps (\cref{fig: maps_gallory}); (ii) A feature detector that is pre-trained on full images to address the interlocking problem, allowing it to identify discriminative features in masked-out regions and break the interlocking cycle; (iii) Our extensive experiments demonstrate that our model outperforms various baselines in predictive accuracy and attribution map quality.




\section{Related Work}
Attribution methods explain deep vision models by identifying input regions that are important to the model’s decision. Existing attribution methods can be categorized into Post-hoc methods and Inherently Explainable methods. 

\noindent\textbf{Post-hoc Methods.} In post-hoc methods, the model is pre-trained and fixed. Post-hoc method can be generally divided into three categories. The first category is activation-based methods \cite{zhou2016learning,selvaraju2017grad,jalwana2021cameras,jiang2021layercam,ramaswamy2020ablation}, which rely on the activation values of the internal neurons to explain the prediction. The first work \cite{zhou2016learning} named Class Activation Mapping (CAM), averages the output of the final convolutional layer (activations) to generate the attribution map. Score-CAM \cite{wang2020score} uses activations as masks and averages layer-wise class activations with classification scores. Ablation-CAM \cite{ramaswamy2020ablation} generates attribution maps by ablating (removing) feature map units in the final convolutional layer and measuring the drop in class activation scores. The second category is backpropagation-based methods~\cite{smilkov2017smoothgrad,srinivas2019full,yang2023local,lundstrom2022rigorous,selvaraju2017grad}, which assumes that the importance of inputs is often measured by the gradients to inputs. However, gradients are found to be independent of the discriminative models and can not serve as valid explanations \cite{srinivasrethinking}. The third category is perturbation-based methods \cite{petsiuk2018rise,dabkowski2017real,stalder2022you,ribeiro2016should}, which identify the relevant features by removing (or maintaining) parts of the input features and summarizing changes in classification scores. RISE \cite{petsiuk2018rise} generates attribution maps by randomly masking the input image and computing the weighted sum of these masks. To save inference time, a selector is trained to generate attribution maps. The selector is trained to preserve the classification score on the masked-in features of the image \cite{dabkowski2017real}. Recently, the selector is further modified to have class-wise output and better fitted into multi-label prediction scenario \cite{stalder2022you}.


\noindent\textbf{Inherently Explainable Methods.} Inherently explainable methods train the selector and predictor jointly \cite{jethani2021have,ganjdanesh2022interpretations,chen2018learning,yoon2018invase}. Specifically, the selector generates attribution maps that select discriminative features. The selected features are then fed into a predictor to conduct predictions. Compared with post-hoc methods, they can generate more faithful attribution maps because the predictor only uses selected features to conduct predictions. However, as mentioned in the introduction, degenerate cases will happen because of the interlocking between the selector and the predictor \cite{jethani2021have,ganjdanesh2022interpretations}. To prevent the selector mislead the predictor, REAL-X \cite{jethani2021have} trains the predictor model with randomly masked images. Then, it fixes the predictor and trains the selector by maximizing the classification score. RB-AEM \cite{ganjdanesh2022interpretations} argues that random masks ignore the geometric prior where nearby pixels share similar semantic information. As a result, RB-AEM trains the predictor with random Radial Basis Function style masks. These methods can mitigate degenerate cases by isolating the predictor training process from the selector. However, there is a distribution shift where the features selected by the selector differ from randomly selected ones. The predictor which is trained on randomly masked images would have sub-optimum performance when tested on features selected by the selector. \textit{Different from these approaches, our proposed approach can avoid degenerate cases while training the predictor and the selector jointly, maintaining high prediction performance}.

\begin{figure*}[t]
\centering
\includegraphics[width=\textwidth]{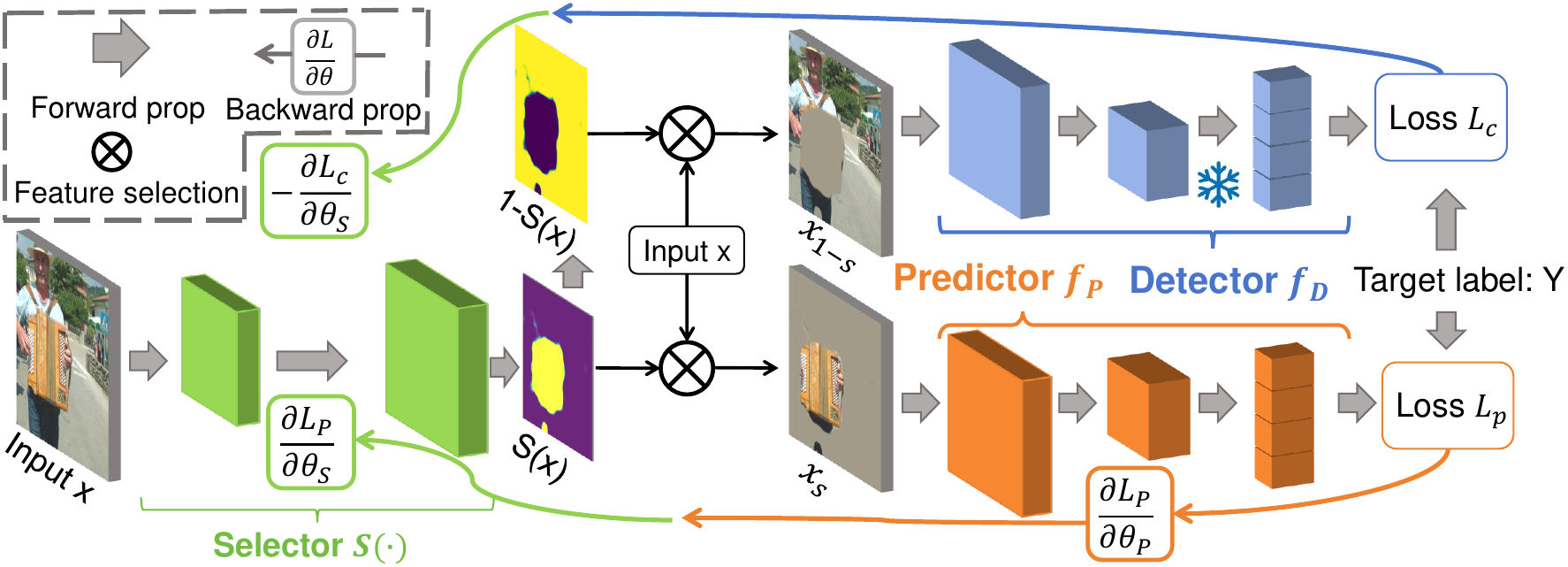}
\caption{The framework consists of a selector $S$, predictor $f_P$, and a pre-trained detector $f_D$. The selector first identifies discriminative features and generates the attribution map $S(x)$. The $S(x)$ selects features $x_s$, which are then fed into the predictor. Meanwhile, the remaining parts of the image, represented as $x_{1-s}$, are fed into the feature detector. The selector is optimized to assist the predictor in accurately predicting the label while confusing the detector by not leaving any discriminative features in $x_{1-s}$.}
\label{fig:framework}
\end{figure*}

\section{The Proposed Framework: {\mymodel}}
We propose a novel framework, {\mymodel}, as illustrated in \cref{fig:framework}. Our framework comprises three modules: the selector $S$, the predictor $f_P$, and the detector $f_C$. 
The selector $S$ identifies discriminative features and generates the attribution map, which is used to select important features for explainable prediction. These selected features are then fed into the predictor for label prediction. The remaining parts of the image are fed into a feature detector introduced to detect discriminative features in the masked-out regions. The detector is pre-trained on full images, enabling it to detect discriminative features without being influenced by the selector. The selector is optimized to assist the predictor in accurately predicting the label while confusing the detector by ensuring that masked-out regions do not contain discriminative information.


We focus on the image classification task. 
Let $x \in \mathbb{R}^{H\times W \times C}$ be an input image with $H$, $W$, and $C$ being the height, width, and number of channels. Let $y \in [1, 2, 3, ...K] $ be the ground-truth label of $x$ with $K$-being the number of classes. We represent the attribution map as $S(x) \in [0, 1]^{H\times W}$ where $S(\cdot)$ is the feature selector and each element of $S(x)$ represents the importance of the corresponding pixel. The attribution map is used as a mask to select discriminative features. Specifically, the masked-in features $x_s$ and masked-out features $x_{1-s}$ are represented as:
\begin{align}
    x_s = S(x) \odot x + (1 - S(x)) \odot q, \quad 
    x_{1-s} = (1 - S(x))  \odot x + S(x) \odot q,
\end{align}
where $q$ is the matrix of the average value of pixels in the dataset. $x_s$ is the masked-in features and $x_{1-s}$ is the masked-out features.

\subsection{The Predictor and Feature Detector}

Since we want to build an inherently explainable model, different from the post-hoc explanation method, the predictor is trained with masked-in features so that it can conduct prediction tasks cooperatively with the selector. Specifically, we train the predictor with the following loss function:
\begin{align}
\label{P_obj}
     \operatorname*{\min}_{P(\cdot)} \mathcal{L}_P(x_s) = H(p(y|x),f_P(x_s)),
\end{align} 
where $H(\cdot,\cdot)$ is the cross-entropy and $p(y|x)$ is the ground-truth label. The predictor is optimal when correctly predicting the target with selected features. 


Next, considering the interlocking between the selector and the predictor, our method introduces a feature detector as an independent check. Trained with full images, the detector remains unaffected by the selector's decisions. We train the detector with the following objective:
\begin{align}\operatorname*{\min}_{D(\cdot)} \mathcal{L}_D(x) = H(p(y|x),f_D(x)).
\end{align} 
The detector is optimal when it can accurately predict the target label. Training on full images $x$ allows the detector to observe features beyond those selected by the selector $x_s$ (same for $x_{1-s}$). This is crucial when the selector's choices are sub-optimal or unreliable. If the selector masks out important features and the predictor fits on the noise, the predictor itself can not effectively detect the discriminative features in masked-out regions. At this time, the detector is needed to identify any discriminative features in the masked-out regions and break the interlocking between the selector and the predictor by penalizing the selector. We will show how the detector can break the interlocking between the selector and the predictor in the following section.

\subsection{Optimizing the Feature Selector}
With the predictor and the detector, we then introduce how to optimize the selector. The goal of the selector is to generate the attribution map. We first define what is an ideal attribution map. Then, we demonstrate how our training framework optimizes the selector to generate this ideal attribution map.

\begin{definition} (Ideal Attribution Map).
For every input image and label $(x, y)$, an optimal attribution map $S(x)$ should separate discriminative features (signal) and non-discriminative features (noise). There are two requirements for the ideal attribution map:
\begin{itemize}
    \item Sufficiency: $x_s$ contains sufficient discriminative features: $H(y|x_s)=H(y|x)$, where $H(y|x_s)$ is the conditional entropy of the label $y$ given the selected discriminative features $x_s$. It requires the label information provided by the selected features $x_s$ is equivalent to that provided by the entire image.    
    \item Completeness: $x_{1-s}$ only contains non-discriminative features: $H(y|x_{1-s}) = H(y)$. This means that the label information provided by the masked-out regions is no different from a random guess.  
\end{itemize}
\end{definition}

We propose the attribution generation as a signal-noise decomposition process. A good attribution map should not only contain enough features for label prediction (sufficiency) but also try to include all discriminative features (completeness). With the definition of ideal attribution maps, we then introduce how our learning framework can optimize the selector to achieve the sufficiency and completeness requirements. We train the selector with the following objective:
\begin{align}\operatorname*{\min}_{S(\cdot)} \mathcal{L}_S(x) = \mathcal{L}_P(x_s) - a*\mathcal{L}_D(x_{1-s}),
\end{align}
where the first term is to help the predictor predict the label and the second term is to discourage the detector from predicting the label accurately for the masked-out regions. 

\textbf{Interlocking Mitigation.} The second term is crucial for addressing the interlocking problem. When the selector selects noise, leading the predictor to fit on this noise, the system reaches a local optimum with a low $\mathcal{L}_P(x_s)$. In this scenario, the detector, which is already trained with full images, can observe and identify the discriminative features that the selector has overlooked. The detector can approximate the label based on discriminative features in masked-out regions, even though it has not been explicitly trained on masked images. As a result, $\mathcal{L}_D(x_{1-s})$ will be low, leading to a high selector loss $\mathcal{L}_S(x)$. This high loss signals the need for the selector to adjust its choices, encouraging it to escape the local optimum and select more discriminative features, thereby breaking the interlocking problem and aligning with the requirements of sufficiency and completeness. With certain assumptions, we then prove the optimal selector can satisfy these requirements.

\begin{proposition} (Optimal Selector).
Assume that the predictor and the detector can effectively predict the labels of their input, i.e., $f_P(x_s) = p(y|x_s)$ and $f_D(x_{1-s}) = p(y|x_{1-s})$, then the optimal selector satisfies both the sufficiency and completeness requirements.
\end{proposition}

\noindent\textbf{Remark.} The $p(y|x_{s})$ and $p(y|x_{1-s})$ refer to the label given the masked-in ($x_{s}$) and masked-out ($x_{1-s}$) regions of the image. There are three situations. (i) Optimal Attribution Map: All discriminative features are selected for $x_{s}$ and only noise is left for $x_{1-s}$. In this case, the label of masked-in regions is the same as the full image (i.e., $p(y|x_{s})=p(y|x)$) and the label of the masked-out regions should be a uniform distribution (i.e., $p(y|x_{1-s}) = p(y)$). This situation means all discriminative features are selected and the attribution map is already ideal. (ii) Worst Attribution Map: Only noise is selected and discriminative features are all in masked-out regions. As discussed in the previous paragraph, the detector can approximate the label based on the discriminative features, resulting in low $\mathcal{L}_D(x_{1-s})$. This increases the selector loss $\mathcal{L}_S(x)$, which compels the selector to include discriminative features. Therefore, our loss function inherently discourages the attribution maps to only include noise, ensuring this scenario is rarely encountered. (iii) Partial Feature Selection: Discriminative features exist both in masked-in and masked-out regions (e.g., A dog's head is included in $x_s$ and its body is $x_{1-s}$). The labels for both masked-in and masked-out regions are the same as full images (i.e., $p(y|x_{s}) = p(y|x_{1-s}) = p(y|x)$). In this situation, $f_P(x_s) = p(y|x_s)$ holds when predictor is optimized with \cref{P_obj}. As introduced in the previous section, the detector is trained with full images. With the understanding of the complete images, the detector can approximate the label based on discriminative features in masked-out regions (i.e., $f_D(x_{1-s}) = p(y|x_{1-s})$). 


\begin{proof}
Since the predictor and detector can effectively predict the labels with their inputs:
\begin{align}
     \label{S_1} \mathcal{L}_P(x_s)&=H(p(y|x_s),f_P(x_s))=H(y|x_s) \geq  H(y|x), \\
     \label{S_2}  -\mathcal{L}_D(x_{1-s}) &= -H(p(y|x_{1-s}),f_D(x_{1-s})) = -H(y|x_{1-s}) \geq -H(y) .
\end{align} 
Thus, $\mathcal{L}_S \geq H(y|x) -a*H(y)$. The equality can be reached when $ S(x) = S^*(x)$, where $S^{*}(x)$ is the ground-truth mask. $\mathcal{L}_S$ achieves equality if and only if both (\cref{S_1}) and (\cref{S_2}) reach their equalities. Thus, the sufficiency ($H(y|x_s)=H(y|x)$) and the completeness ($H(y|x_{1-s}) = H(y)$) requirements can be satisfied when the selector is optimized.

\end{proof}

\subsection{Final Objective Function of {\mymodel}}
We demonstrate that our framework can generate sufficient and complete attribution maps while making predictions. In practice, a regularizer $R(S(x))$ is added to the selector loss: $\mathcal{L}_S + b*R(S(x))$ to control the sparsity of the attribution map and avoid the trivial example: $x_s = x$. The overall loss for our training framework:
\begin{align}
     \operatorname*{\min}_{S(\cdot),f_P(\cdot)} \mathcal{L}(x) = \mathcal{L}_P(x_s) - a*\mathcal{L}_D(x_{1-s}) + b*R(S(x)).
\end{align}
Directly using the L1 norm as the regularizer is not optimal. Instead, we use a threshold $t$ to control the sparsity of the mask: $R(S(x)) = max( \frac{1}{HW} \sum{S(x)}-t,0)$. If over $t$ fraction of the pixels are selected, the regularizer will penalize the model. The first term of the loss function is to predict the target label via the collaborative work of the selector and predictor. This term ensures that the attribution map contains enough information about the target (Sufficiency). The second objective term is to penalize the selector if the detector $f_C$ can infer the target label with masked-out regions. This term discourages the presence of discriminative features in the masked-out regions (Completeness). An overall training algorithm is in the supplementary materials.


\noindent\textbf{Discussion} Recently, several other methods have been trying to mitigate the interlocking between the selector and the predictor. Instead of using the selected features $x_s$, they train the predictor with randomly masked images $x_{s^{\prime}}$, either uniformly at random ($s^{\prime} \sim \text{ Bernoulli(0.5)}$) as suggested by \cite{jethani2021have} or using a Radial Basis Function (RBF)-style masking ($s^{\prime} \sim \text{ Bernoulli}(\text{RBF})$) as proposed by \cite{ganjdanesh2022interpretations}. Then, they fix the predictor while training the selector. These approaches mitigate the interlocking problem by isolating the training process of the predictor from the selector. However, these mitigation approaches are sub-optimal because the predictor is trained with $x_{s^{\prime}}$ (randomly selected features) which still differs from the distribution of $x_s$ (selector selected features). The distribution shift would lead to low prediction accuracy and a less precise attribution map. 

Learning with three modules has been employed in various tasks, such as domain adaptation \cite{ganin2015unsupervised,zhang2017aspect}, learning sleep stages radio signals \cite{zhao2017learning}, and rationalization in text data \cite{yu2019rethinking}. These works use an adversarial framework to learn three modules, aiming to obtain domain-invariant representations or rationales of the input text. In contrast, our work focuses on improving the attribution maps of deep vision models. Additionally, unlike the adversarial frameworks in previous studies, our feature detector is pre-trained on full images and remains fixed during the training of the other two modules. Training the detector on masked-out regions would cause it to fit the noise in these areas, reducing its effectiveness in detecting discriminative features. By pre-training on complete images, the detector maintains its ability to see the entire picture without being misled by noise. To ensure the detector provides reliable signals from the start of the training process, it is pre-trained and kept fixed.


\section{Empirical Validation}

In this section, we conduct experiments to answer the following research questions: (RQ1) Can our proposed model conduct accurate predictions? (RQ2) Can our proposed model produce high-quality attribution maps?

\subsection{Experimental Setup}
\textbf{Datasets.} We conduct experiments on two benchmark datasets: (i) \textbf{ImageNet-9} \cite{xiao2020noise}: It is a subset of ImageNet containing 9 classes. For each sample in the test set, ImageNet-9 also has pixel-wise masks that highlight the target object;  (ii) \textbf{NICO++} \cite{zhang2023nico++}: NICO++ dataset contains 60 classes. The dataset has 11313, 4440, and 4440 samples for training, validation, and testing. (iii) The an8Flower\cite{oramasvisual} dataset has predefined discriminative regions of a synthetic plant model. Each class is distinguished by assigning one of six colors to either a flower or a stem, while the remaining part defaults to green. The detailed information about the datasets is in supplementary materials.


\noindent\textbf{Baselines} We compare {\mymodel} with representative and state-of-the-art explanation methods, including three post-hoc explanation methods, i.e., Grad-CAM \cite{selvaraju2017grad},  RT-Saliency \cite{dabkowski2017real}, and Score-CAM \cite{wang2020score}, and four inherently explainable methods, i.e., B-cos \cite{bohle2022b}, REAL-X \cite{jethani2021have}, RB-AEM \cite{ganjdanesh2022interpretations}, and DiET (Distractor Erasure Tuning) \cite{bhalla2023discriminative}. We further summarize these approaches and a detailed description of the baselines is in supplementary materials.

\noindent\textbf{Implementations.} We use Resnet18 \cite{he2016deep} for both the predictor and detector. To control the overall size of our model, we use the Lite R-ASPP Network model \cite{sandler2018mobilenetv2} with MobileNetV3 \cite{howard2019searching} as the backbone for our selector. All methods are trained from scratch for a fair evaluation and to avoid the influence of external knowledge. Images are normalized with the mean and standard deviation of the datasets. For post-hoc baselines, we apply them on the Resnet18. For inherently explainable baselines, we use Resnet18 as the backbone. We conduct a hyper-parameter search for all the methods using the validation set. For our model, the optimal coefficients are found $a=5$, $b=100$, and $t=0.1$. The learning rate is set as $0.0005$. The maximum epoch is set as $200$ for all methods.


\subsection{RQ1: Assessing Classification Accuracies}
To answer RQ1, we assess the classification accuracy of our model in comparison to regular Resnet and inherently explainable baselines. Furthermore, to examine the robustness of the models, we evaluate models on the noisy version of these two datasets where we randomly remove 10\% of pixels in each image. We also test \mymodel-TD which is a variant of our model where the detector is trained on masked-out regions along with the selector and the predictor. We conduct experiments 5 times and the average results with standard deviation. In \cref{tab:Prediction}, we can observe that our method outperforms other baselines on ImageNet-9 and demonstrates greater improvements on NICO++. Our model is also robust on the noisy version maintaining the highest accuracy among baselines. Other inherently explainable methods have a lower accuracy compared with regular Resnet18 and Resnet101. The reason is that they either have extra constraints that limit the prediction ability (B-cos) or would potentially discard discriminative features that can be helpful for prediction (REAL-X, RB-AEM, DiET). Our model achieves higher accuracy than Resnet18 and Resnet101. There are two possible reasons. One is that our detector module encourages all discriminative features to be seen by the predictor ensuring that our module would not perform worse than regular Resnet. Moreover, removing non-discriminative noise can improve the overall prediction performance \cite{rao2023using}.



\begin{table}[t]
\small
    \centering
    \caption{Classification Performance (Accuracy$\pm$Std \%)}
    \begin{tabular}{l r r r r}
          \toprule
          
        Model   & Imagenet9 & NICO++ & Imagenet9(noisy) & NICO++(noisy) \\
        
      \midrule

      Resnet18 \cite{he2016deep} &  91.16 \scriptsize{$\pm$0.34} &  40.20 \scriptsize{$\pm$0.45}   & 66.53 \scriptsize{$\pm$1.13} & 28.97 \scriptsize{$\pm$0.44} \\
      Resnet101 \cite{he2016deep} & 91.96 \scriptsize{$\pm$0.06} & 39.58 \scriptsize{$\pm$0.40} & 64.91 \scriptsize{$\pm$1.32} & 28.64 \scriptsize{$\pm$1.24} \\
      B-cos \cite{bohle2022b} & 75.21 \scriptsize{$\pm$0.70} & 27.04 \scriptsize{$\pm$1.13} & 59.75 \scriptsize{$\pm$0.63} & 22.72  \scriptsize{$\pm$0.69}\\
      Realx \cite{jethani2021have} & 66.13 \scriptsize{$\pm$2.55} &21.11 \scriptsize{$\pm$0.11}& 53.97 \scriptsize{$\pm$0.76} & 8.63 \scriptsize{$\pm$1.24}\\
      RB-AEM \cite{ganjdanesh2022interpretations} & 78.40 \scriptsize{$\pm$0.34} &17.93  \scriptsize{$\pm$0.44}& 50.30 \scriptsize{$\pm$0.98} & 6.84 \scriptsize{$\pm$1.29} \\
      DiET \cite{bhalla2023discriminative} & 73.41 \scriptsize{$\pm$1.21} & 38.45 \scriptsize{$\pm$0.07} &54.71  \scriptsize{$\pm$0.75} & 7.93 \scriptsize{$\pm$0.33} \\ \hdashline
     \mymodel-TD & 92.82 \scriptsize{$\pm$1.07} &  \textbf{58.38} \scriptsize{$\pm$0.87} & 70.32 \scriptsize{$\pm$0.85} &  \textbf{38.30} \scriptsize{$\pm$0.17} \\ 
      \mymodel & \textbf{94.33} \scriptsize{$\pm$0.37} & 56.47 \scriptsize{$\pm$1.74} & \textbf{74.81} \scriptsize{$\pm$0.93} &  33.12 \scriptsize{$\pm$0.10}\\
      
      \bottomrule

    \end{tabular}
    
    \label{tab:Prediction}
\end{table}

\begin{table}[ht]
        \centering
        \caption{Area under the IOU curve (The higher the better)}
        \resizebox{\linewidth}{!}{
        \begin{tabular}{cccccccccc}
            \toprule
             Grad-CAM & Score-CAM & RT-Saliency & Bcos & DiET & COMET & COMET(ViT) & COMET-DR \\
            \midrule
             15.74 & 17.51 & 25.3 & 39.2 & 11.6 & \textbf{72.5} & 69.5 & 5.83 \\
            \bottomrule
        \end{tabular}
        }
        \label{tab:iou}
\end{table}

\begin{table}[ht]
\centering
\caption{Localization ability test: PxAP of different methods. (The higher the better)  }

\resizebox{\textwidth}{!}{
\begin{tabular}{ l c c | ccccccc}
\hline
\multicolumn{3}{c}{Post-hoc} & \multicolumn{6}{c}{Inherent} \\
\hline
Grad-CAM & Score-CAM & RT-Saliency & Realx & RB-AEM & B-cos & DiET & \mymodel-TD & \mymodel & \mymodel(ViT) \\
\hline
51.15  & 53.55   & 60.67   & 44.05   & 40.61   & 60.36  & 47.68 & 71.41 & \textbf{76.38} &69.87 \\
\hline
\end{tabular}}
\label{table:localization}
\end{table}

\subsection{RQ2: Assessing the Quality of Attribution Maps}

To answer RQ2, we evaluate the qualities of feature attributions. We quantify the feature coverage, localization ability, fidelity, and robustness of our model.

\textbf{Feature Coverage:} To assess whether our models can accurately cover discriminative features, we calculate the area under the IOU (intersection over union) curve for the an8Flower, which has predefined discriminative regions \cite{oramasvisual}. Additionally, we perform a sanity check \cite{adebayo2018sanity} by training \mymodel-DR on randomized data, where labels are randomly assigned to images. As shown in \cref{tab:iou}, our models achieve the highest scores, indicating that they accurately cover the discriminative features. In contrast, \mymodel-DR, which scores the lowest, fails to generate accurate attribution maps. This significant difference from \mymodel demonstrates that our models produce valid attribution maps closely related to the training data.

\textbf{Localization Ability:} 
We use PxAP (Pixel Average Precision) \cite{choe2020evaluating} to evaluate attribution maps pixel-by-pixel against ground-truth maps and calculate the area under the precision-recall curve. The detailed definition of PxAP is in supplementary materials. As shown in \cref{table:localization}, our proposed method outperforms other baselines. Among post-hoc methods, RT-Saliency outperforms activation-based methods (Grad-CAM and Score-CAM). The reason is that the relevance of neuron activation to the prediction is often not clear and can be heuristic. RT-Saliency directly measures the impact of altering specific features on the prediction. Realx and RB-AEM have overall low localization ability because the predictor is trained on randomly masked images that are not in the same distribution as the ones from the selector. Compared with \mymodel-TD, the \mymodel~ has better performance. \mymodel-TD would also be unstable because the detector would fit on noise in the masked-out regions, which degrades its ability to detect discriminative features. Without the influence of the selector, the pre-trained detector can effectively detect discriminative features in masked-out regions.

\textbf{Fidelity:} Fidelity refers to the accuracy of the attribution method in correctly identifying the features that are truly important or influential in the model's decision-making process. As a result, pixel perturbation test \cite{choe2020evaluating,srinivas2019full,bhalla2023discriminative} is used to quantify the fidelity of our attribution method. We mask k\% of pixels with the highest attribution scores, where k is varied as \{5, 10, 20\}. We then test the masked images with the predictor and record the accuracy. Low accuracy indicates that the masked pixels are important for prediction. Results are shown in \cref{table:fidelity}. Inherently explainable methods will produce lower accuracy when pixels with high attribution scores are perturbed. This means that inherently explainable methods produce more faithful attribution maps, which is also shown in the previous study \cite{jethani2021have}. Our method outperforms other baselines and correctly identifies important features.

\textbf{Robustness:} For attributions, we also evaluate the robustness of our model. We randomly remove 20\% pixels and calculate the PxAP of generated attribution maps. We test on the same test set where only the backgrounds of images are changed. \cref{tab:Robustness} shows that our method has the best performance with both noisy images and background-changed images.



\begin{table}[t]
\small
    \centering
    \caption{Fidelity test: Perturbed accuracy when removing pixels with top attribution scores. (Accuracy$\pm$Std \%, the lower the better)}
    \resizebox{\textwidth}{!}{
    \begin{tabular}{c | c | r r r | r r r}
        \toprule
       \multirow{2}{*}{Type} & \multirow{2}{*}{Model} &  \multicolumn{3}{c}{ImageNet-9} & \multicolumn{3}{c}{NICO++} \\ 
       \cline{3-8}
        & & top 5\% & top 10\% & top 20\% & top 5\% & top 10\% & top 20\%\\
      \hline 
      \multirow{3}{*}{Post-hoc} & Grad-CAM \cite{selvaraju2017grad}   & 85.20 \scriptsize{$\pm$0.16}& 79.93 \scriptsize{$\pm$0.42} & 69.96  \scriptsize{$\pm$0.30} & 29.27 \scriptsize{$\pm$0.91} & 22.20 \scriptsize{$\pm$0.40} & 14.57 \scriptsize{$\pm$0.28}  \\
      & Score-CAM \cite{wang2020score} & 84.46 \scriptsize{$\pm$0.58} & 78.63 \scriptsize{$\pm$0.98}  & 67.23 \scriptsize{$\pm$1.04} & 28.02 \scriptsize{$\pm$0.03}  & 21.54 \scriptsize{$\pm$0.27} & 13.84 \scriptsize{$\pm$0.17}   \\
      & RT-Saliency \cite{dabkowski2017real} & 83.91 \scriptsize{$\pm$0.04}  & 75.26 \scriptsize{$\pm$1.13}   & 59.12 \scriptsize{$\pm$4.14} & 34.02 \scriptsize{$\pm$0.59} & 30.32 \scriptsize{$\pm$0.16} & 24.84 \scriptsize{$\pm$0.30} \\
      \hline 
      
      \multirow{6}{*}{Inherent} & Realx \cite{jethani2021have} &25.14 \scriptsize{$\pm$2.23} &23.70 \scriptsize{$\pm$1.53}  &21.48 \scriptsize{$\pm$1.22} & 11.20
\scriptsize{$\pm$1.94} & 10.02
\scriptsize{$\pm$1.96}  & 4.41
\scriptsize{$\pm$1.26} \\
      & RB-AEM \cite{ganjdanesh2022interpretations} &67.76 \scriptsize{$\pm$4.12} &64.53 \scriptsize{$\pm$4.86}   &56.79 \scriptsize{$\pm$3.73}  & 11.92 \scriptsize{$\pm$0.85} & 11.04 \scriptsize{$\pm$0.31}  & 6.24 \scriptsize{$\pm$0.24} \\
      &B-cos \cite{bohle2022b}  &53.24 \scriptsize{$\pm$0.68} &42.74 \scriptsize{$\pm$0.41}  &31.43 \scriptsize{$\pm$0.70} & 14.38 \scriptsize{$\pm$0.17} & 10.39 \scriptsize{$\pm$0.06} & 7.00 \scriptsize{$\pm$0.08} \\
      &DiET \cite{bhalla2023discriminative} & 27.94 \scriptsize{$\pm$0.12} & 26.01 \scriptsize{$\pm$0.01}  &23.57 \scriptsize{$\pm$0.51} & 20.22 \scriptsize{$\pm$0.53} & 12.61 \scriptsize{$\pm$0.05} & 6.64 \scriptsize{$\pm$0.18}\\ \hdashline
      &\mymodel-TD & 16.34 \scriptsize{$\pm$0.72} &  15.90 \scriptsize{$\pm$0.71}  & 14.23 \scriptsize{$\pm$0.87} & \textbf{6.11} \scriptsize{$\pm$0.18} & \textbf{2.99} \scriptsize{$\pm$0.24} & \textbf{1.87} \scriptsize{$\pm$0.16}\\ 
      &\mymodel & \textbf{10.64} \scriptsize{$\pm$1.12} & \textbf{8.68} \scriptsize{$\pm$1.25} & \textbf{2.07} \scriptsize{$\pm$1.24}  & 8.97 \scriptsize{$\pm$0.13} & 4.69 \scriptsize{$\pm$0.27} & 2.74 \scriptsize{$\pm$0.49}  \\
      \bottomrule
    \end{tabular}}
    \label{table:fidelity}
\end{table}

\begin{table}[t]
\small
    \centering
    \caption{Robustness test: PxAP when randomly removing pixels or changing backgrounds (the higher the better).}
    \begin{tabular}{l l r r r}
        \toprule
       \multicolumn{2}{l}{Model}    & Noisy images  & Background change\\
        \midrule
      \multirow{3}{*}{Post-hoc} & Grad-CAM \cite{selvaraju2017grad} &  47.28 \scriptsize{$\pm$0.98}   & 52.54 \scriptsize{$\pm$0.52}  \\
      & Score-CAM \cite{wang2020score} & 52.90 \scriptsize{$\pm$0.92} & 53.14 \scriptsize{$\pm$0.85} \\
      & RT-Saliency   \cite{dabkowski2017real}      & 57.53 \scriptsize{$\pm$1.19} & 54.21 \scriptsize{$\pm$3.06}\\
      \midrule
      \multirow{6}{*}{Inherent} & Realx \cite{jethani2021have} & 40.93  \scriptsize{$\pm$0.97}   &42.52 \scriptsize{$\pm$0.53}  \\
      & RB-AEM \cite{ganjdanesh2022interpretations} & 38.91 \scriptsize{$\pm$0.89}  &39.54 \scriptsize{$\pm$0.41}  \\
      &B-cos \cite{bohle2022b} & 57.28 \scriptsize{$\pm$0.67}  & 57.85 \scriptsize{$\pm$0.59}\\
      &DiET \cite{bhalla2023discriminative}  & 45.20 \scriptsize{$\pm$1.35}  & 42.25 \scriptsize{$\pm$1.23} \\ \hdashline
      &\mymodel-TD & 70.30 \scriptsize{$\pm$1.60}  & 73.33 \scriptsize{$\pm$0.56} \\ 
      &\mymodel & \textbf{73.04} \scriptsize{$\pm$0.85} & \textbf{74.22} \scriptsize{$\pm$1.15}\\
      \bottomrule
    
    \end{tabular}
    \label{tab:Robustness}
\end{table}

\subsection{Further Assessment: Ablation and Hyper-Parameter Sensitivity}

Aside from comparing our model with baselines, we conduct an ablation study to show the effectiveness of the detector: (i) \textbf{\mymodel-TD (Trainable Detector):} We train the detector with masked-out regions instead of pre-training it with full images; (ii) \textbf{\mymodel-w/o Detector}: We remove the detector and optimize both the selector and the predictor. In this setting, the predictor takes over the task of the detector detecting any discriminative features in masked-out regions; (iii) \textbf{\mymodel-FP (Fixed Predictor)~\& w/o Detector}: In this setting, the detector is removed and the predictor is also pre-trained and fixed. We only optimize the selector.

    

\begin{figure*}[t]
\centering
\includegraphics[width=\textwidth]{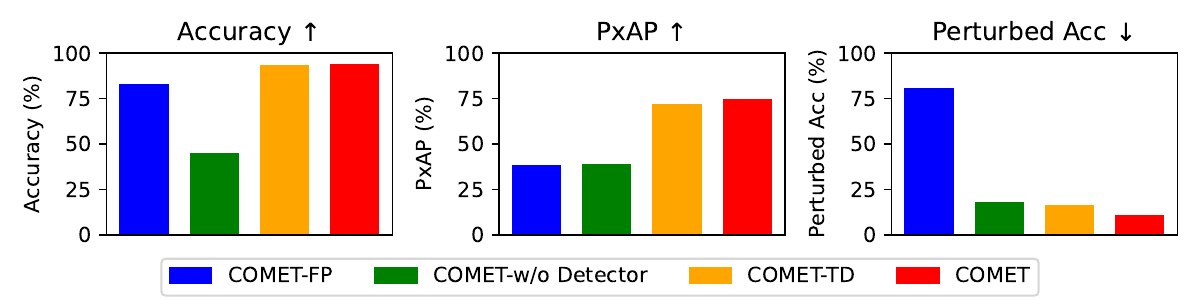}
\caption{Ablation study.}
\label{fig:Ablation}
\end{figure*}

\begin{figure*}[t]
\centering
\includegraphics[width=.24\textwidth]{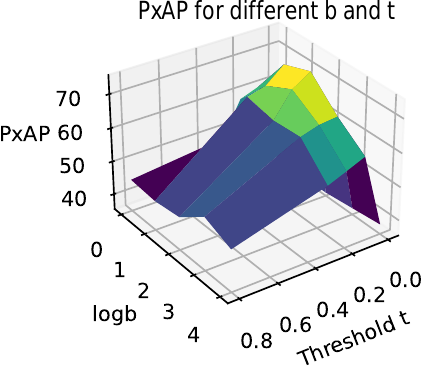}\hfill
\includegraphics[width=.24\textwidth]{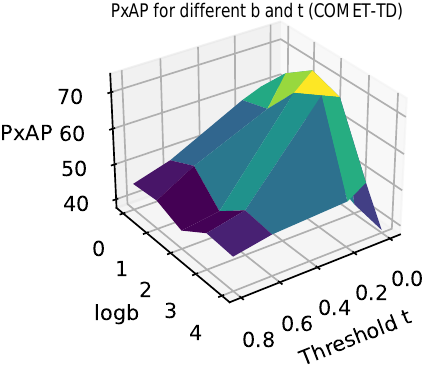}\hfill
\includegraphics[width=.24\textwidth]{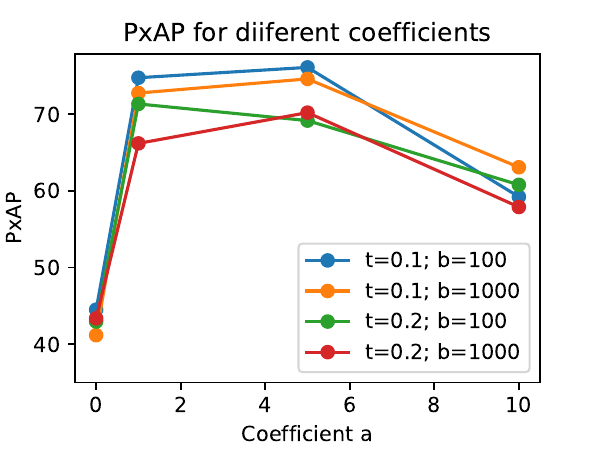}\hfill
\includegraphics[width=.24\textwidth]{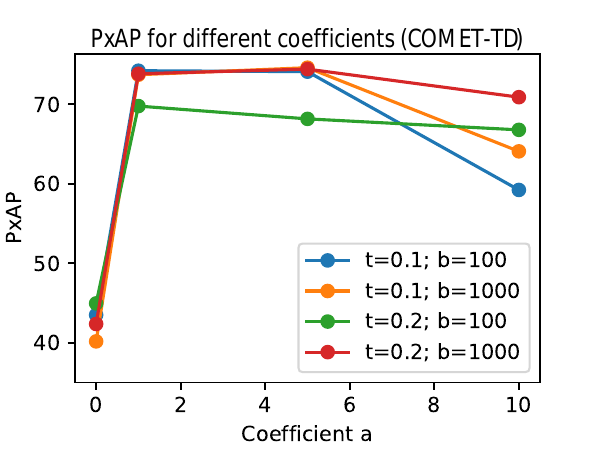}\hfill
\caption{Test for different coefficients. The left two figures are PxAP for different $b$ and $t$. The right two figures are PxAP for different $a$.}
\label{fig:coef_test}
\end{figure*}

        

      
      

In \cref{fig:Ablation}, all other models have overall lower perturbed accuracy than \mymodel-FP, because training the predictor and the selector jointly will have higher fidelity and find truly important pixels for the predictor. By comparing \mymodel-FP and \mymodel-w/o Detector, we can see serious degenerate cases where the accuracy drops quickly. Considering that \mymodel-w/o Detector makes the predictor trainable, the model should have obtained higher accuracy with more weights. However, the selector and predictor interlock each other, leading to even lower prediction accuracy. \mymodel~improves the prediction accuracy and localization ability (PxAP) of attribution maps. \mymodel~has better results with \mymodel-TD, because a pre-trained detector can effectively detect discriminative features and make the training process more stable. We perform the hyper-parameter sensitivity analysis by trying different values of $a$, $b$, and $t$. We test both \mymodel~and \mymodel-TD. Firstly, we try different values of $b$ and $t$, which control the mask regularization term. $b$ controls stength the mask regularization term the overall training process. $t$ is a threshold value and larger $t$ indicates more pixels can be accessed by the predictor. As shown in \cref{fig:coef_test}, the PxAP of both \mymodel~and \mymodel-TD reaches the top values when $b \in \{100,1000\}$ and $t \in \{0.1,0.2\}$. We further test different values of $a$. \mymodel~reaches the maximum value when $a=5$, $b=100$ and $t=0.1$. The PxAP decreases dramatically when $a=0$ (No control on masked-out regions). This shows that the control on masked-out regions can improve the attribution maps.





\subsection{Visualization}
\begin{figure}[t]
\centering
\includegraphics[width=\linewidth]{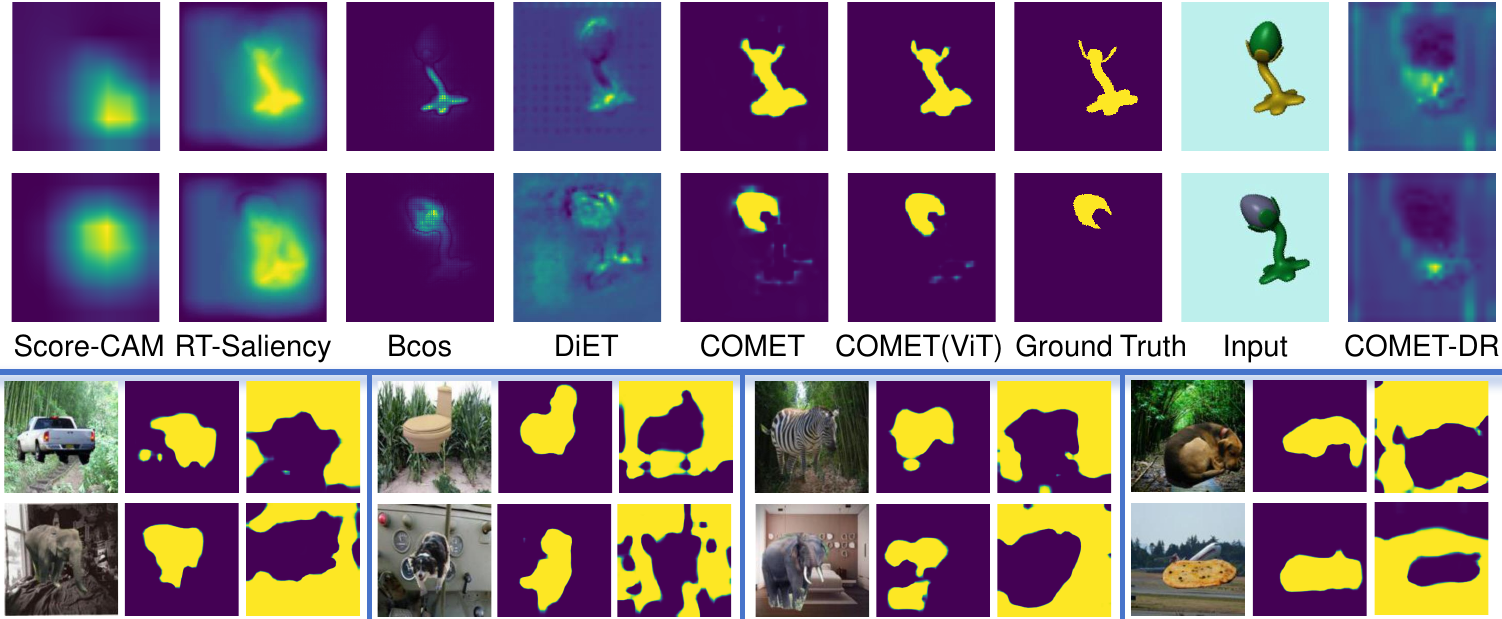}
\caption{Top (an8Flower): Attribution maps comparison; Bottom (BAM): Attribution maps when model trained on the object (left) or scene (right) labels.}
\label{fig:flower_examples}
\end{figure}

As shown in \cref{fig:flower_examples}, \mymodel~highlights the color-assigned, discriminative part of the plant while excluding the non-discriminative (green) part. This shows that our model can accurately cover discriminative features instead of highlighting the whole object. We also train our model on the BAM dataset, synthesizing objects and scenes \cite{BAM2019}. Using the same training set, we train two models: one with object labels and another with scene labels. As shown in \cref{fig:flower_examples} (bottom), the object-labeled model focuses on the object, while the scene-labeled model emphasizes the scenes.

\section{Discussion}

Despite the effectiveness, we recognize two limitations. The sparsity of the mask can be challenging to uniformly apply due to varying target sizes across images. Secondly, the model's complexity leads to longer training time. However, this is a one-time investment, and our model has the shortest inference time. We compare computation time and report model size in the supplementary materials.


\section{Conclusion}
Our proposed model \mymodel, effectively addresses the incompleteness problem by introducing an additional objective that penalizes the presence of discriminative features in the masked-out regions. This enhances the comprehensiveness of feature selection, resulting in accurate and complete attribution maps. An independent detector is introduced to detect discriminative features in masked-out regions. The independent detector can effectively break the interlocking situation of the selector and the predictor. Experiments demonstrate that our method can generate high-quality attribution maps while making accurate predictions.

\section*{Acknowledgements}

This work  was in part supported by the National Science Foundation (NSF) awards \#1934782 and \#2114824, the Army
Research Office (ARO) under grant number W911NF-21-1- 0198, Department of Homeland Security (DHS) CINA under grant number
E205949D, and Cisco Faculty Research Award.

%
%
\bibliographystyle{splncs04}
\bibliography{main}
\end{document}


\title{Supplementary Materials} 

\titlerunning{Comprehensive Attribution for Vision Models}

\author{Xianren Zhang\inst{1}\orcidlink{0000-0003-0283-4693} \and
Dongwon Lee\inst{1}\orcidlink{0000-0001-8371-7629} \and
Suhang Wang \inst{1}\orcidlink{0000-0003-3448-4878}}

\authorrunning{X. Zhang et al.}


\institute{The Pennsylvania State University, University Park, PA, USA \\
\email{\{xzz5508,dongwon,szw494\}@psu.edu}}

\maketitle

\section{Applications} 

\begin{figure*}[htbp]
\centering
\includegraphics[width=\textwidth]{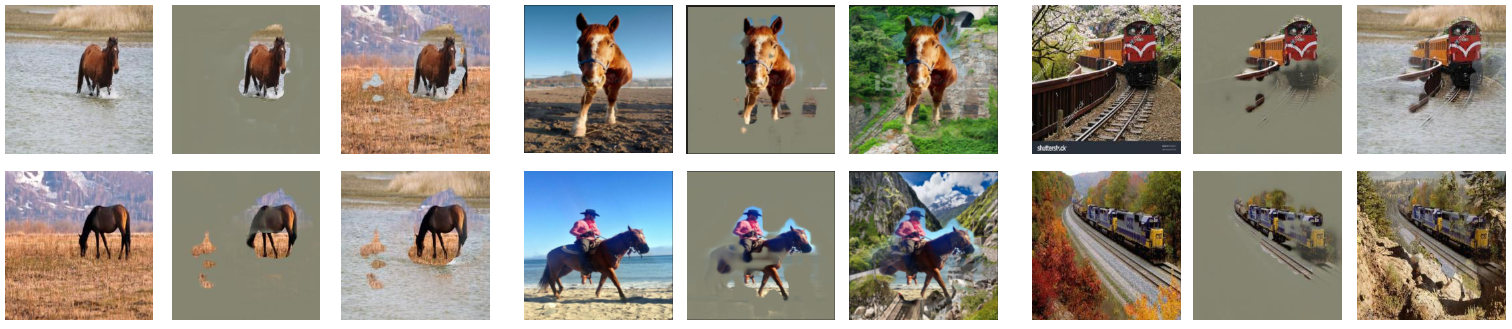}
\caption{Counterfactual samples generation}
\label{fig:Counterfactual_examples}
\end{figure*}

There are three main potential applications for our proposed model, as follows:
\begin{itemize}
    \item \textbf{Trustworthiness:} Our method can enhance the trustworthiness of machine learning models in critical domains such as healthcare. By providing comprehensive explanations for the model's decisions, especially in the analysis of medical images, doctors can make better decisions with a higher level of confidence. 

    \item \textbf{Model debugging:} Since our model can provide attribution maps with high fidelity, it can also be applied in model debugging \cite{adebayo2020debugging}. One can check if a model focuses on spurious features such as the backgrounds of images via attribution maps. Moreover, supervision over attribution maps can guide the model and improve its performance \cite{rao2023using}. 

    \item 
    \textbf{Image Generation: } Some works try to remove spurious correlation between an object and its background by generating more counterfactual samples where the same objects appear in different backgrounds \cite{wu23disc}. However, they directly mix up the original image and the background image, making the generated images unrealistic and unnatural. They also need an external generation model \cite{rombach2022high} to get more backgrounds. On the other hand, our model, as illustrated in Figure \ref{fig:Counterfactual_examples}, can generate natural counterfactual samples where the same target appears in a different background with the help of our precise attribution map. We can fully utilize the dataset itself to generate more counterfactual images.
\end{itemize}

\section{Algorithm}

The details of the training algorithm for COMET are illustrated in \cref{alg: algorithm1}. The training algorithm involves three steps. First, it initializes the selector, predictor, and detector modules. Then, it pretrains the detector to accurately predict the target labels with full images. Finally, the algorithm focuses on training the selector and predictor jointly. The selector generates an attribution map to highlight important features in the image, while the predictor uses these features to make predictions. During this process, the detector penalizes the selector if it masks out important features, guiding the selector to include all relevant information in the attribution map.

\begin{algorithm}
\caption{Training Algorithm for \mymodel}
\label{alg: algorithm1}
\textbf{Input:} Dataset: $\mathcal{D}$;\\
\textbf{Output:} Selector $S(\cdot)$ and predictor $f_P(\cdot)$;\\
\textbf{Select:} Learning rate $\alpha$, Hyper-parameters: $a$, $b$ and $t$.
\begin{algorithmic}[1]
\STATE Initialize the selector $S(\cdot)$, predictor $f_P(\cdot)$, and feature detector $f_D(\cdot)$.

\FOR{$1,...,T$}
\FOR{$(x,y) \in \mathcal{D}$}
\STATE Compute the loss $\mathcal{L}_D(x) = -y\log f_D(x)$.
\STATE $ \theta_{D} = \theta_{D} - \alpha \frac{\partial \mathcal{L}_{D}(x)}{\partial\theta_{D}}$.
\ENDFOR
\ENDFOR

\FOR{$1,...,T$}
    \FOR{$(x,y) \in \mathcal{D}$}
    \STATE Generate attribution map $S(x)$.
    \STATE The masked-in image $x_s = S(x) \odot x + q \odot (1-S(x))$.
    \STATE The masked-out image $x_{1-s} = (1 - S(x)) \odot x + q \odot S(x) $.
    \STATE Compute the predictor loss $\mathcal{L}_P(x_s) = -y\log f_P(x_s)$.
    \STATE Compute the detector loss $\mathcal{L}_D(x_{1-s}) = -y\log f_D(x_{1-s})$.
    \STATE Compute the regularizer $R(S(x))= max( \frac{1}{HW} \sum{S(x)}-t,0)$.
    \STATE Overall loss $\mathcal{L}(x) = \mathcal{L}_P(x_s) - a \cdot \mathcal{L}_C(x_{1-s}) + b \cdot R(S(x))$.
    \STATE $ \theta_{P} = \theta_{P} - \alpha\frac{\partial \mathcal{L}(x)}{\partial\theta_{P}}$ and $ \theta_{S} = \theta_{S} - \alpha \frac{\partial \mathcal{L}(x)}{\partial\theta_{S}}$.
    \ENDFOR
\ENDFOR
\end{algorithmic}
\end{algorithm}

\section{Details of Datasets}

Below, we provide more details on the two datasets used in our evaluation.

\begin{itemize}
    
\item \textbf{ImageNet-9} \cite{xiao2020noise}: It is a subset of ImageNet containing 9 classes (e.g., dog, bird, fish, monkey and vehicle). It has 45405, 4185, and 4050 samples for training, validation, and testing respectively. The validation set has 4185. The test set has 4050 samples (450 samples for each class). ImageNet-9 also provides images whose backgrounds are switched with other images. We use background-changed images to test the robustness of our model. For each sample in the test set, ImageNet-9 has a pixel-wise ground-truth attribution map that highlights the target object. 

\item \textbf{NICO++:} \cite{zhang2023nico++}: NICO++ dataset is a large-scale domain generalization benchmark. It contains 60 classes (e.g., horse, cow, train, truck, and tiger) in 10 common domains (e.g., rock, dim, water, and autumn). Since there is no split into training, validation, and test sets. We split the dataset randomly, with 11313, 4440, and 4440 samples for training, validation, and testing. 
\end{itemize}

\section{Baseline Details}
\begin{itemize}
\item Grad-CAM \cite{selvaraju2017grad}: Grad-CAM is a representative method in activation-based approaches. It aggregates the neuron activations and gradient information to generate attribution maps.

\item RT-Saliency \cite{dabkowski2017real}: RT-Saliency is a perturbation-based method. This method trains an external generator by maximizing the classification score on the masked-in features. It generates the attribution maps in real-time with a single forward pass.

\item Score-CAM \cite{wang2020score}: Score-CAM is a combination of the activation method and perturbation method. Score-CAM uses the activation maps as masks and gets the classification scores of the model with masked input. To generate the attribution map, Score-CAM uses classification scores as coefficients of activation maps, instead of the final layer's weights \cite{zhou2016learning} or gradients \cite{selvaraju2017grad}.

\item B-cos \cite{bohle2022b}: B-cos network is a recently proposed inherently explainable model. It converts the complex non-linear network as an input-dependent linear transform where the output can be viewed as the sum of input weighted by input-dependent weights.

\item REAL-X \cite{jethani2021have}: Cooperatively training the selector and the predictor will have degenerate cases. To overcome this problem, REAL-X trains the predictor model separately with random masks. Then, it fixes the predictor and trains the generator by maximizing the classification score.

\item RB-AEM \cite{ganjdanesh2022interpretations}: RB-AEM (Radial based amortized explanation model) argues that REAL-X neglects geometric prior and Random Radial Basis Function (RBF)-like functions are employed to train the predictor.

\item DiET (Distractor Erasure Tuning) \cite{bhalla2023discriminative}: DiET masks out unimportant features with iterative mask rounding with increasing sparsity. This iterative rounding process can remove the noise step by step. 

\end{itemize}

\section{PxAP definition}
The precision and recall at threshold $\tau$ are defined as the following:
\begin{align}
     \label{precision} Prec(\tau) = \frac{| \{ S(x)_{ij} \geq   \tau \} \cap \{ GT_{ij} = 1 \} |}{| \{ S(x)_{ij} \geq \tau \} |}, \\
     \label{recall} Rec(\tau) = \frac{| \{ S(x)_{ij} \geq  \tau \} \cap \{ GT_{ij} = 1 \} |}{| \{ GT_{ij} = 1 \} |},
\end{align} 
where the $GT$ is the ground truth mask. We can calculate precision and recall with different thresholds. PxAP is the area under the precision-recall curve: $PxAP = \sum_{k=1}^{K} (Rec(\tau_{k}) - Rec(\tau_{k-1})) \times Prec(\tau_k)$. $\tau$ is $20$ in our experiment.

\section{ Running Time}

We test the computation time of generating an attribution map for each method, as shown in \cref{tab:training_time}. For each method, we generate attribution maps for 100 samples and average the consumed time for one sample. Score-CAM takes much more time than other methods because it has multiple candidate masks. It has to run multiple times to score these masks and then aggregate them to generate the final attribution maps. In comparison, we can observe that our method and RT-Saliency consume the least time to generate attribution maps because they only need one forward pass of the selector to generate maps. 

For training time, CAM-based methods (Grad-CAM and Score-CAM) take the least time because they only need the predictor (resnet18 in our experiment) to be trained. However, they suffer from low-fidelity problems and can not effectively identify really important features for the predictor as shown in fidelity test. Our method does not have an advantage in training time because we have three modules in our framework. It's important to note that training is a one-time investment, but the payoff is a model that is inherently interpretable and can provide high-quality attribution maps with the least inference time.

\begin{table}
\small
\centering
\caption{Computation time}
    \begin{tabular}{l r r r r r r r}
    \multicolumn{8}{@{}c}{\em(a) Inference time}\\
        \toprule
        Model & Grad-CAM & Score-CAM & RT-Saliency & Realx & RB-AEM & B-cos & \mymodel \\
        \midrule
        Time (s) & 0.025 & 1.054 & \textbf{0.021} & 0.029 & 0.031 & 0.028 & \textbf{0.021} \\
        \bottomrule
    \end{tabular}
    \label{tab:inference_time}

\vspace{1em}

    \begin{tabular}{l r r r r r r }
    \multicolumn{7}{@{}c}{\em(b) Training time for one epoch}\\
        \toprule
        Model & Resnet18 (CAM) & RT-Saliency & Realx & RB-AEM & B-cos & \mymodel \\
        \midrule
        Time (s)  & \textbf{66.81} & 209.50 & 176.96 & 168.52 & 142.50 & 193.05 \\
        \bottomrule
    \end{tabular}
    \label{tab:training_time}

\label{tab:time}
\end{table}

\section{Visualization}

\begin{figure*}[htbp]
\centering
\includegraphics[width=\textwidth]{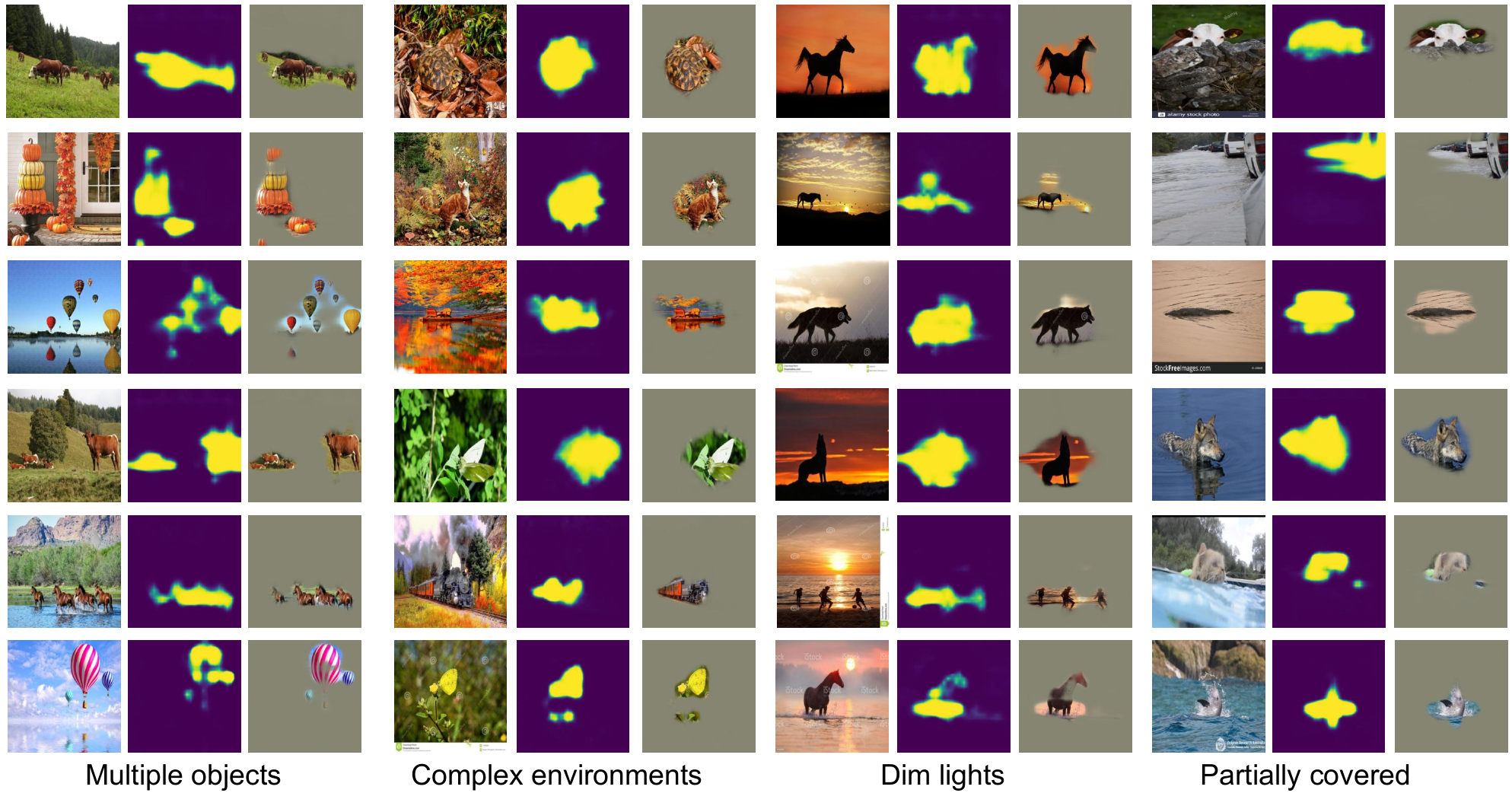}
\caption{Attribution maps of different scenarios.}
\label{fig:nico_examples_supp}
\end{figure*}

In \cref{fig:nico_examples_supp}, we display various attribution maps generated by our method under various challenging visual scenarios. These maps effectively demonstrate the model's robust feature localization capabilities. \textbf{Multiple Objects:} The model successfully identifies and highlights multiple targets within a single image, indicating the comprehensiveness of our attribution method. \textbf{Complex Environments:} Even with distracting backgrounds (e.g., similar colors), the model can isolate the main subjects from the backgrounds. \textbf{Dim Lights:} The attribution maps maintain their precision in dim conditions. \textbf{Partially Covered:} The model exhibits the capacity to highlight objects that are not fully visible, despite the loss of details and features.

\section{Accuracy vs Model Size}

In this section, we evaluate the trade-off between model size and accuracy on ImageNet-9 as shown in \cref{tab:model_selection}. We compare various combinations of selectors and predictors in terms of their accuracy and model size. Also, we report the size and accuracy of different layers of Resnets. Notably, the combination of LRASPP as the selector and Resnet18 as the predictor strikes a balance between accuracy and model size. This combination achieves high accuracy with a relatively small model size. Consequently, we choose this combination as our final model. It is worth mentioning that our model outperforms both Resnet34 and Resnet101. Despite their larger sizes, these models achieve lower accuracy.

\begin{table}
\small
    \centering
    \caption{Accuracy and size for different models (Selector + Predictor)}
    \begin{tabular}{c c c}
      \hline
     Models & Acc (\%) & Size (M) \\ \hline
        LRASPP+Res18 (\mymodel)   & 93.70   & 14.399 \\ 
      DeepLabV3+Res18   & 93.80   & 22.202   \\ 
     LRASPP+Res34  & 93.60 & 24.508 \\
       FCN+Res18 & 91.14 & 44.128 \\
        Resnet18  & 91.16  &  11.181 \\
        Resnet34  & 91.14 &21.315\\
        Resnet101  & 91.96  & 42.519 \\
       \hline

    \end{tabular}
    \label{tab:model_selection}
\end{table}

\section{Extended Experiments on Robustness}
We also use the original attribution maps, generated by our model with clean images, as a reference, while keeping the setup in the paper unchanged. We then calculate the PxAP to check if the attribution maps are affected by the noise or background. As shown in \cref{tab:robustness}, our method has the highest PxAP, which demonstrates the robustness of the model.

\begin{table}
\small
    \centering
    \caption{PxAP with original attribution maps as ground truth}

\begin{tabular}{ccc}
\hline
Models & Noisy images & Background change \\
\hline

Grad-CAM & 70.77 & 73.80 \\
Score-CAM & 68.22 & 72.79 \\
RT-Saliency & 66.32 & 54.98 \\
Realx & 60.61 & 55.27 \\
RB-AEM & 74.48 & 71.60\\
Bcos & 69.46 & 53.63 \\
DiET & 63.47 & 36.23 \\
COMET & \textbf{77.93}  &  \textbf{77.54} \\

\hline
\end{tabular}

\label{tab:robustness}

\end{table}

\section{Limitations and Future work}

Despite the effectiveness of our model, we acknowledge two limitations. 
First, the sparsity of the mask is controlled by the regularization term, $R(S(x))$, along with parameters $b$ and $t$, which can be challenging to uniformly apply due to varying target sizes across images. An intuitive way to mitigate this problem would be to allow the model to autonomously determine the number of pixels to include. Yet, this approach does not work, as the model tends to include more pixels to minimize loss, often incorporating irrelevant background features. Additionally, due to the complexity of the model, our method needs sizable time in training, as shown in \cref{tab:training_time}. However, note that such a training time is a one-time investment while our model consumes the least time in inference, which is critical for a practical solution. 

There are several potential topics for future study. Since our model conducts selections and then predictions, the robustness of our model to adversarial attacks and spurious correlation is a promising topic. Since the model conducts selection operation, it is also possible that attacks and spurious correlation in  background can be discarded. Study in natural language models finds that selecting important words for prediction would be helpful to avoid the influence of spurious correlation \cite{ross2022does}. Additionally, human evaluation on attribution maps can also be further explored. Our model can further be modified to have class-wise output, which can provide more information and fit into multi-class prediction task.

%
%
\bibliographystyle{splncs04}
\bibliography{main}